\documentclass[letterpaper, 10 pt, conference]{ieeeconf}  %

\IEEEoverridecommandlockouts                              %
\usepackage[utf8]{inputenc} %
\usepackage[T1]{fontenc}    %
\usepackage{url}            %
\usepackage{booktabs}       %
\usepackage{amsfonts}       %
\usepackage{nicefrac}       %
\usepackage{microtype}      %
\usepackage{lipsum}
\usepackage{graphicx}
\usepackage[caption=false,font=footnotesize]{subfig}

\usepackage{graphics} %
\usepackage{epsfig} %
\usepackage{mathptmx} %
\usepackage{times} %
\usepackage{amssymb}  %
\usepackage{xcolor}
\usepackage{float}
\usepackage{amsmath}
\usepackage{siunitx}
\usepackage{textcomp, gensymb}
\usepackage{comment}
\usepackage{mathrsfs}

\usepackage[bookmarks=true]{hyperref}
\title{\LARGE \bf
Constrained Reinforcement Learning for Safe Heat Pump Control%
}

\author{Baohe Zhang\authorrefmark{1}\authorrefmark{2}, Lilli Frison\authorrefmark{1}\authorrefmark{4}, Thomas Brox\authorrefmark{2}, Joschka Bödecker\authorrefmark{2}%
\thanks{\authorrefmark{1}: These authors contributed equally}%
\thanks{\authorrefmark{2}: Faculty of Computer Science,
        University of Freiburg, Germany. \authorrefmark{4}: Systems Control and Optimization Laboratory, Department of Microsystems Engineering,
        University of Freiburg, Germany. Contact Address: {\tt\small zhangb@cs.uni-freiburg.de}}
}

\begin{document}
\maketitle
\thispagestyle{empty}
\pagestyle{empty}

\begin{abstract}
We study heat pump control in buildings as a constrained reinforcement learning (RL) problem: minimize electrical energy while keeping indoor temperature within comfort bounds. We formulate the task as a constrained Markov decision process with temperature comfort constraints, argue adn verify that optimal operation typically lies near the comfort boundary and that a smoothed log‑barrier variant of Soft Actor‑Critic (CSAC‑LB) exploits this structure, and report robustness to sensor noise and model mismatch. We release I4B, a lightweight simulator with a Gym‑style API and a built‑in MPC baseline to enable reproducible RL studies. On two building scenarios, CSAC‑LB balances comfort and energy at least on par with MPC while maintaining constraint satisfaction. Benchmarking against several baseline algorithms demonstrates CSAC-LB's efficiency in exploration and control performance.
\end{abstract}

\section{Introduction}
\label{sec:introduction}
Heat pumps (HP) have become the central component of modern building heating systems, offering high efficiency but requiring precise control to balance comfort and energy use. Constrained RL methods can balance thermal comfort with energy usage while learning from noisy sensor data in real time, as demonstrated in robotic tasks \cite{DBLP:conf/rss/HaarnojaHZTTL19}. They also enable continuous learning in dynamic scenarios where building usage or electricity prices change.

Simulation frameworks play a vital role in training RL agents. %
An accurate, adaptable, and parallelizable simulator can significantly speed up training, leading to more precise control over heating systems.
Existing frameworks for heating control lack a comprehensive, open-source, and lightweight simulator that combines multiple building environments, flexible customization and explicit HP actuator control. To fill this gap, we propose Intelligence for Building (I4B), a novel open-source framework for advanced HP control strategies like model predictive control (MPC) and RL for heat pump operation. I4B provides an interface between the building simulation module and control algorithms, incorporating reference controllers, support for parallelization, and standardized metrics for evaluation.

We frame the heating control problem as a constrained Markov Decision Process (CMDP), aiming to minimize energy usage while maintaining indoor temperature above a set threshold. By applying state-of-the-art constrained RL algorithms within this framework, we benchmark their performance across various scenarios. Notably, the constrained RL with linear smoothed log barrier function (CSAC-LB) proves particularly suitable for heating control problems characterized by optimal solutions at the boundary of feasible and infeasible sets. Our findings show that CSAC-LB excels in balancing exploration and performance. Our contributions are as follows:
\begin{itemize}
    \item We propose I4B\footnote{Publicly available under \url{https://github.com/lfrison/i4b}.}, a new open-source lightweight building heat pump operation simulator with rich customization options and interfaces for different research communities.
    \item We apply a variety of constrained RL algorithms to different heating scenarios. An empirical study is performed to benchmark them. Code and experiments are publicly available.
    \item We demonstrate that CSAC-LB can balance the objective and constraints better compared to other SOTA methods, while achieving efficient exploration.
\end{itemize}
\section{Related Work} \label{sec:related_work}
\subsection{RL in Building Heating Control}
\label{sec:RL_in_HVAC_related_work}
RL has emerged as a promising approach for optimizing HVAC systems, achieving energy savings of 5–12\% over standard controls~\cite{brandi2020deep, nagy2023ten}. Research focuses on comparing model-free RL with strategies like MPC, model-based RL, and hybrid methods~\cite{nagy2018deep, rohrer2023deep}. For instance, \cite{ARROYO2022118346} found that pure RL struggled with state constraints, leading to reinforced predictive control (RL-MPC) that combines RL and MPC strengths. Learning-based MPC also utilizes neural networks for system modelling~\cite{pmlr-v242-frison24a}. Despite progress, further exploration of advanced RL algorithms, particularly constrained RL for constraint satisfaction, is needed in building heating control.

\subsection{Constrained RL Algorithms} \label{sec:safe_rl_related_work}
Significant advancements in constrained RL are summarized in recent reviews \cite{DBLP:journals/jmlr/GarciaF15,DBLP:journals/ml/Dulac-ArnoldLML21}. Safe policy search methods integrate nonlinear programming into policy gradients, including gradient projection techniques \cite{4354030, DBLP:conf/icml/Bou-AmmarTE15, DBLP:conf/nips/YangJDZZL0022}, with extensions using Gaussian Processes for risk estimation \cite{polymenakos2019safe}. Constrained Policy Optimization (CPO) \cite{DBLP:conf/icml/AchiamHTA17} and its successors \cite{DBLP:conf/nips/0010VR20} employ trust-region methods with theoretical guarantees. CVaR optimization has also been applied \cite{DBLP:conf/nips/ChowTMP15,DBLP:journals/jmlr/ChowGJP17}. Extensions of SAC \cite{DBLP:conf/icml/HaarnojaZAL18} incorporate cost functions and Lagrange multipliers \cite{DBLP:conf/corl/HaXTLT20, ray2019benchmarking}, with further improvements using distributional safety critics \cite{DBLP:conf/aaai/YangSTS21}. Interior-point methods have been applied to on-policy RL algorithms \cite{DBLP:journals/corr/abs-1812-06502,DBLP:conf/aaai/LiuDL20}. Model-based approaches similar to MPC, like \cite{DBLP:conf/nips/BerkenkampTS017}, use dynamics models to ensure safety but are limited by model accuracy assumptions.

\subsection{Building Simulators} \label{sec:building_simulator_related_work}

Several environments wrap high-fidelity engines such as EnergyPlus or Modelica and expose them to RL via FMI/FMUs (Functional Mock-up Interface/Units): Energym~\cite{Energym}, BOPTEST~\cite{BOPTEST}, Sinergym~\cite{Sinergym}, COBS~\cite{COBS2020}, and FlexDRL~\cite{FlexDRL}. These frameworks are well suited for building energy studies and, depending on the test case or configuration, can be used for HP control. In our summary (Table.~\ref{tab:compare_envs}), Energym and several BOPTEST test cases expose HP-level actuators (e.g., compressor frequency or supply-water setpoint), whereas Sinergym/COBS/FlexDRL are HP-capable when the underlying IDF/Modelica setup includes the required actuators. Typical action spaces in these wrappers are EnergyPlus/Modelica actuators or zone setpoints; vectorized training is uncommon because simulations run as separate external processes.
Beobench~\cite{Beobench} is an experiment orchestrator that aggregates backends. HP-capability depends entirely on the chosen backend and configuration, and actions/observations are passed through without defining a fixed plant model.
More lightweight simulators avoid external engines. Google SBsim~\cite{GoldfederSipple2023BuildSys} provides an internal building simulator aimed at large-scale experimentation; by default it focuses on device/zone schedules and does not expose HP-level actuators. SustainGym~\cite{SustainGymNeurIPS2023} uses reduced-order RC (resistance–capacitance) models and targets building energy optimization via zone-level HVAC power, which is practical for benchmarking but not the same as direct HP control. Our I4B uses RC models as well, but exposes HP-level actions (supply-water temperature setpoint or compressor fraction), includes auto generators for weather and internal gains, and ships with a built-in MPC baseline, making HP control experiments straightforward (Table.~\ref{tab:compare_envs}).

\section{Constrained RL}
\subsection{Heat pump operation as a CMDP}
We formulate heat-pump control as a CMDP $\langle S,A,P,r,\gamma,c\rangle$ by discretizing the task with sampling time $\Delta t$. State:$s_t=[T_{\mathrm{room}},\,T_{\mathrm{wall}},\,T_{\mathrm{ret}},\,T_{\mathrm{amb}},\,q_{\mathrm{solar}},\,q_{\mathrm{int}}]$,
where $T_{\mathrm{room}}$, $T_{\mathrm{wall}}$, $T_{\mathrm{ret}}$, $T_{\mathrm{amb}}$ are temperatures ($^\circ\mathrm{C}$) for indoor air, a wall node,\footnote{In hardware, $T_{\mathrm{wall}}$ is typically unmeasured; deployment is then Partial Observable MDP-like with a Kalman filter over $(T_{\mathrm{room}},T_{\mathrm{amb}},T_{\mathrm{sup}},T_{\mathrm{ret}})$. In simulation we expose full state for a clean benchmark.}
HP return water, and ambient air; $q_{\mathrm{solar}},q_{\mathrm{int}}$ are solar/internal heat gains (W).
Action: $a_t=T^{\mathrm{set}}_{\mathrm{sup}}$ is the HP supply-water setpoint temperature ($^\circ\mathrm{C}$).
Dynamics $P$: with step $\Delta t$, the next state is given by the discretized thermal model from Eq.~\ref{eq:dynamics0} and Eq.~\ref{eq:dynamics}:
\[
s_{t+1}=f_\Delta\!\big(s_t,a_t;\,T_{\mathrm{amb},t},q_{\mathrm{solar},t},q_{\mathrm{int},t}\big).
\]
Reward function (negative energy per step) is:
$$
r_t=-P_{\mathrm{el}}(s_t,a_t)\,\Delta t,\qquad
P_{\mathrm{el}}(s_t,a_t)=\frac{\dot Q_{\mathrm{hp}}(s_t,a_t)}{\mathrm{COP}(T^{\mathrm{set}}_{\mathrm{sup}},T_{\mathrm{amb},t})},
$$
where $\mathrm{COP}(\cdot)$ is a standard second-order fit to manufacturer data. The constraint cost (comfort deviation) is defined as:
\[
c_t=\max\!\bigl(T_{\mathrm{ref}}-T_{\mathrm{room},t},\,0\bigr),\qquad T_{\mathrm{ref}}=20\,^\circ\mathrm{C},
\]
so $c_t$ (K) is the instantaneous shortfall below the comfort bound. Over a fixed horizon of $N$ steps, we constrain the discounted sum of deviations
\[
J_{\mathrm{c}}(\pi)\;=\;\mathbb{E}_\pi\!\Big[\sum_{t=0}^{N-1} \gamma^{t} c_t\Big]\;\le\; c_{\max},
\]
where $c_{\max}$ is the maximum allowed sum of the temperature deviation. With discount factor $\gamma$, the CMDP thus reads
\[
\pi^\star=\arg\max_{\pi}\; \mathbb{E}_\pi\!\Big[\sum\nolimits_{t=0}^{N-1} \gamma^{t} r_t\Big]\;
\text{s.t.}\; J_{\mathrm{c}}(\pi)\le 0,
\]
which anchors all baselines (RL and MPC) to the same objective and constraint.

\subsection{Linear Smoothed Log Barrier Function}
The log barrier method is recognized for its capability to address constrained optimization problems that include inequality constraints, yet it is prone to challenges with numerical stability. This instability arises because the logarithmic function within the log barrier method cannot accommodate conditions where $g(x) > 0$, leading to difficulties in maintaining stable optimization.

In the realm of deep RL, where neural networks are used to approximate value functions and model continuous-control policies, ensuring constraint satisfaction is particularly challenging. Due to the randomness of neural network initialization, the policy defined by the actor network may initially violate constraints, yielding an unsafe policy. A straightforward workaround with log-barrier penalties is to clip the constraint-violation cost to avoid numerical instability near feasibility. However, such clipping can introduce saturation and flat gradients during optimization, impeding learning and potentially masking the severity of violations.

To solve these drawbacks associated with directly clipping the neural network's output, a novel approach involving a linear smoothed log barrier function~\cite{logbarrierExtension}, $\psi(x)$, has been introduced. This function is as follows:
\begin{align}
\label{eq:ext_log_barrier}
\psi(x) = \begin{cases}
       - \frac{1}{\mu} \log (-x) & \text{if $x \leq -\frac{1}{\mu^{2}}$}\\
      \mu x - \frac{1}{\mu}\log (\frac{1}{\mu^{2}}) + \frac{1}{\mu} & \text{otherwise}\\
    \end{cases}       
\end{align}
where $\mu$ is a parameter that adjusts the function's behavior. The key advantage of $\psi(x)$ lies in its continuous and differentiable nature across its entire domain, allowing for the application of stochastic gradient descent (SGD) without being confined solely to the feasible set. This facilitates a more robust and flexible approach to optimizing neural networks in the context of constrained RL.

\subsection{SAC-Lag}
Originally developed for teaching quadruped robots locomotion~\cite{DBLP:conf/corl/HaXTLT20}, SAC-Lag adapts the Soft Actor-Critic framework~\cite{DBLP:conf/icml/HaarnojaZAL18} by incorporating stepwise constraints to limit the robot's pose, aiming to mitigate potential damage. To define this stepwise constrained optimization task, $d_{t}$ as the permissible threshold for constraint violations at each step $t$, with the primary goal being to maximize the following objective:
\begin{align}
&\sum^{T}_{t=0} \mathop{\mathbb{E}}_{\mathit{a}_t \sim \pi (\mathit{s}_t)} \biggr[ \gamma^{t} \mathit{R}(\mathit{s}_t, \mathit{a}_t) + \alpha \mathit{H}(\pi(\cdot | \mathit{s}_t))\biggr ], \\ & \textrm{s.t.} ~  \mathop{\mathbb{E}}_{\mathit{a}_t \sim \pi (\mathit{s}_t)} \biggr[C(\mathit{a}_{t}, \mathit{s_{t}}) - d_{t} \biggr] \leq 0, ~ \forall t
\end{align}
where $\mathit{R}$ is the reward function, $C$ is the cost function of constraint violation and $\mathit{H}(\pi(\cdot | \mathit{s}_t))$ is the entropy term.
\cite{ray2019benchmarking} extends it to cumulative cost formulation and use a Lagrange-multiplier method. We follow their approach and denote $d$ as the cumulative cost limit. The unconstrained optimization problem can be formulated as:
\begin{align}
\label{eq:dual_sac}
\max_{\pi}\;\min_{\beta \ge 0}\;
\mathcal{L}(\pi,\beta)
\doteq f(\pi) - \beta\, g(\pi),
\end{align}
where 
\begin{align}
&f(\pi)=\mathbb{E}_\pi\!\Big[\sum_{t=0}^{N-1}\gamma^{t}\big(R(s_t,a_t)+\alpha\,H(\pi(\cdot\mid s_t))\big)\Big], \nonumber\\
&g(\pi)=\mathop{\mathbb{E}}_{(s_t, a_t) \sim \pi} \sum^{N-1}_{t=0} \biggr[\gamma^{t} C(a_{t}, s_{t})  \biggr] - d. \nonumber
\end{align}

We denote the reward Q-network with parameters $\theta_{r}$ as $\mathit{Q}_{\theta_{r}}$ and the cost Q-network with parameters $\theta_{c}$ as $\mathit{Q}_{\theta_{c}}$.
To minimize Eq.~\ref{eq:dual_sac}, Lagrange-multiplier $\beta$ is updated according to the following loss:
\begin{align}
\mathit{J}(\beta) = \mathop{\mathbb{E}}_{\mathit{s}_t \sim \mathit{D}, \mathit{a}_t \sim \pi_{\phi} (\mathit{s}_t)} \biggr[ \beta (d - \mathit{Q}_{\theta_{c}}(\mathit{s}_t, \mathit{a}_t))\biggr ]
\end{align}

This adjustment mechanism increases $\beta$ when the cost Q-network's output surpasses the limit $d$, thereby tightening the constraints, and decreases it as the current policy becomes more likely to satisfy these constraints.

Now, assuming a fixed value for $\beta$, the actor loss can be written as:
\begin{align}
    \mathit{J}(\phi) = \mathop{\mathbb{E}}_{\mathit{s}_t \sim \mathit{D}, \mathit{a}_t \sim \pi_{\phi} (\mathit{s}_t)} \biggr[ \alpha \log \pi_{\phi}(\mathit{a}_t | \mathit{s}_t) - \mathit{Q}_{\theta_{r}}(\mathit{s}_t, \mathit{a}_t) \nonumber\\+ \beta \mathit{Q}_{\theta_{c}}(\mathit{s}_t, \mathit{a}_t)\biggr ]
\end{align}
\subsection{CSAC-LB for heat pump operation}\label{sec:csac_lb}
Empirically, energy consumption decreases as the policy lowers the supply setpoint temperature $T^{\mathrm{set}}_{\mathrm{sup}}$ until the indoor temperature $T_{\mathrm{room}}$ approaches the comfort limit. Hence, at the optimum, the comfort constraint is typically active:
\begin{align}
\nabla_\theta J_R(\pi_\theta)
+ \lambda\,\nabla_\theta J_C(\pi_\theta)
= 0, 
\quad \lambda \ge 0,
\label{eq:kkt_condition}
\end{align}
with the complementarity condition $\lambda\,J_C(\pi_\theta)=0$. When constraint violations occur infrequently, standard Lagrangian updates adjust $\lambda$ only upon violations, which may lead to underestimation of the cost critic. The smoothed log-barrier, in contrast, provides a non-zero gradient near the boundary, steering the policy toward informative boundary states without requiring repeated violations.

With the help of the linear smoothed log-barrier function, the constrained RL problem can be optimized directly via stochastic gradient descent. CSAC-LB~\cite{zhang2024constrained} follows the setup of SAC-Lag~\cite{ray2019benchmarking} and employs double-$Q$ critic networks not only for the reward but also for the cost of constraint violation, taking the maximum of the two cost estimates to mitigate underestimation:
\begin{equation}
\label{eq:ext_log_barrier_rl}
\tilde{\psi}(Q_c(s,a))
~\doteq~
\psi\!\left(\max_{i\in\{1,2\}} Q_{c,i}(s,a) - d\right),
\end{equation}
where $Q_{c,1}$ and $Q_{c,2}$ are the cost critics, $d$ denotes the constraint threshold. The smoothed barrier function $\psi(\cdot)$ replaces hard penalty clipping with a differentiable surrogate, yielding continuous gradients near the boundary and allowing both policy and critics to be optimized end-to-end via SGD.

Applying CSAC-LB to heating-system control offers a distinct advantage: due to the nature of the objectives and comfort constraints, the optimal policy typically operates near the boundary between feasible and infeasible regions. This property naturally encourages exploration along the boundary, improving the accuracy of both reward and cost estimation. By leveraging the linear smoothed log-barrier, CSAC-LB promotes focused exploration within informative regions and enhances data efficiency. In contrast, Lagrangian-based methods often induce excessive or uninformative exploration within either the feasible or infeasible set, owing to their sparse violation updates and the biased data distribution in the replay buffer.

\section{I4B: a lightweight, advanced heat pump control simulator}
\label{sec:simulator}
I4B bridges control theory and reinforcement learning by providing fast, customizable building models with explicit access to heat pump actuators and comfort-related constraints. We first outline the simulator design and then describe a representative thermal building and heat pump models used in our experiments.

\subsection{Interfaces and simulator architecture}
I4B is designed to make CMDP‑style heating studies reproducible with minimal setup: a single Python install, a Gym‑style API, vectorized environments for parallel training, automatic generators for weather, solar and internal gains, and a built‑in MPC reference controller. A random building generator based on TABULA enables training‑test splits across buildings.

\begin{figure}[h]
\centering
\includegraphics[width=1\columnwidth]{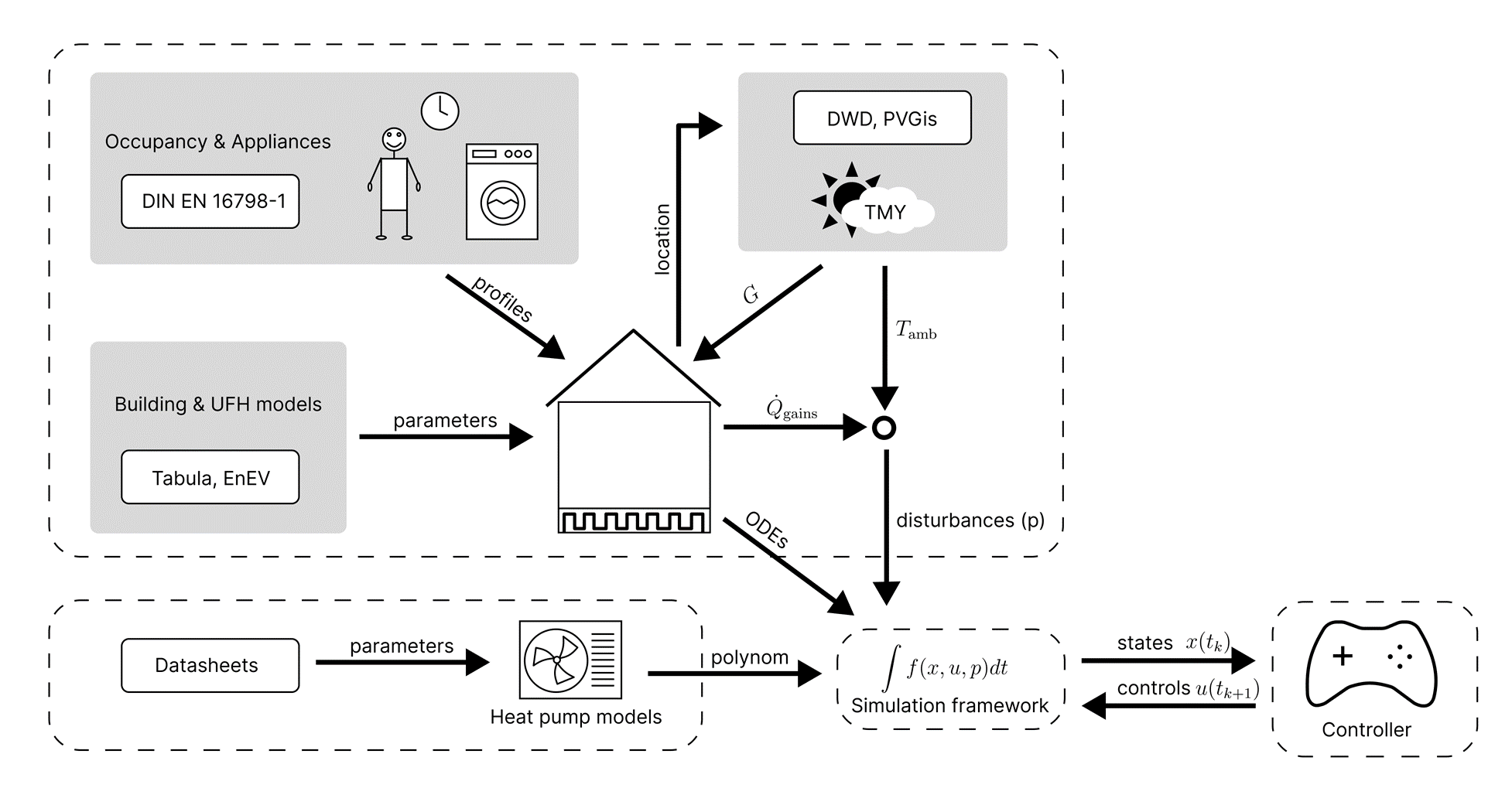}
\caption{Overview of software architecture}\label{fig:i4c-overview}
\end{figure}
The simulator architecture is illustrated in Figure~\ref{fig:i4c-overview}. A simulator class provides a high level interface to perform one and multi-step simulations that return the next state(s) of the building (temperatures) and indicators for comfort levels and the energy demand. This interface can be used to evaluate and test different control strategies, supporting from simple rule-based control policy to RL or MPC-based policy. Disturbance profiles for the ambient temperature, weather forecast, internal heat gains by occupancy, and appliances, as well as solar heat gains, can be generated automatically given the building's location and size. %
The goal of the simulation framework is to quickly generate a reduced-order model of a specific building and simulate this building.
Furthermore, we also support Safety-Gymnasium-style~\cite{DBLP:conf/nips/JiZZP0SGZD023} API for the RL community and it also supports vectorized environment for parallel training.
We have also created tools for randomly generating new building profiles for testing algorithms' generalizability.

\subsection{Modeling choices and positioning of I4B}
Reduced-order RC buildings and single-zone abstractions are standard in building control and HP studies.
TABULA archetypes provide envelope parameters for European buildings \cite{tabula}.
COP polynomial surrogates are common when matching manufacturer data sheets.

Table~\ref{tab:compare_envs} contrasts I4B with popular simulators used in building control studies. Unlike wrappers around large external engines, I4B favors fast iteration for RL while remaining physically grounded via the reduced‑order model. The following simplifications are taken: 
\begin{itemize}
\itemsep0em 
\item Single-zone envelope; no multi-zone air exchange or stratification.
\item Fixed hydronic flow; pump power not modeled explicitly.
\item No humidity or latent loads; air is perfectly mixed.
\item HP defrost cycles and compressor limits are folded into the COP fit bounds.
\item Piping/storage losses aggregated in $k_{\mathrm{UA}}$; domestic hot water excluded.
\end{itemize}

\begin{table*}[t]
\vspace{5pt}
\centering
\caption{Comparison of building-control simulators used in RL papers. “HP-capable” means the environment exposes heat-pump-level actuators (e.g., compressor fraction or supply-water setpoint).}
\label{tab:compare_envs}
\resizebox{1.0\textwidth}{!}{%
\setlength{\tabcolsep}{3pt}
\begin{tabular}{
|c|c|c|c|c|
}%
\toprule
Env. & Ext.Engine & Built-in MPC & HP-capable & Action space (typ.) \\%& Obs./State (typ.) & Key parameters \\
\midrule
Energym~\cite{Energym} & FMI & -- & Yes & HP power fraction \(u\!\in\![0,1]\) \\%& $T_{\mathrm{zone}}$,$T_{\mathrm{amb}}$,$T_{\mathrm{sup/ret}}$,\newline COP,P & E+ weather files, HP type, zone RC \\
BOPTEST~\cite{BOPTEST} & Modelica & -- & Yes & Compressor freq or setpoint (case-dependent) \\% & \(T_{\mathrm{zone}},P_{\mathrm{HP}},T_{\mathrm{sup}},T_{\mathrm{ret}}\) & Testcase config, climate, hydronic HP \\
Sinergym~\cite{Sinergym} & E+ & -- & Possible & Zone setpoints/actuators (IDF/JSON) \\%& Configurable variables/meters & IDF, EPW, actuator config \\
Beobench~\cite{Beobench} & Agg.  & -- & Yes (via backend) & Pass-through to backend env \\% & Backend env Obs. & Selected backend + config \\
FlexDRL~\cite{FlexDRL} & FMU/E+/\newline Modelica & -- & Possible & Subsystem setpoints (single building) \\% & Obs. per co-sim setup & E+/Modelica co-sim, real testbed data \\
COBS~\cite{COBS2020} & E+  & -- & Possible & E+ actuator setpoints (model-dependent) \\%& E+ outputs \newline(zones/devices) \\% & IDF/EPW, building selection \\
Google SBsim\newline\cite{GoldfederSipple2023BuildSys} & No (Int.)  & -- & No (default) & Device setpoints (e.g., supply-water setpoint)\\% & Device/zone temps, occupancy, weather & Building object, ActionConfig, step interval \\
SustainGym~\cite{SustainGymNeurIPS2023} & No (RC)  & Yes & No & Zone HVAC power (per zone) \\%& \(T_{\mathrm{zones}},T_{\mathrm{amb}}\), irradiance, occupancy & RC matrices, weather\\
I4B (ours) & No (RC)  & Yes & Yes & Supply-water setpoint or compressor fraction \\%& $T_{\mathrm{zone}}$,$T_{\mathrm{wall}}$,$T_{\mathrm{ret}}$,$T_{\mathrm{amb}}$,\newline $q_{\mathrm{solar}}$,$q_{\mathrm{int}}$ & building/weather/gain characteristics, step size \\
\bottomrule
\end{tabular}%
}
\vspace{2pt}
{\footnotesize \textbf{Abbrev.:}\; E+=EnergyPlus,\; FMI=Functional Mock-up Interface,\; FMU=Functional Mock-up Unit,\; RC=Resistance--Capacitance Model,\; Int.=Internal simulator,\; Agg.=Aggregator/wrapper.}
\end{table*}

\subsection{Thermal building models}
\begin{figure}
    \centering
    \includegraphics[width=0.5\linewidth]{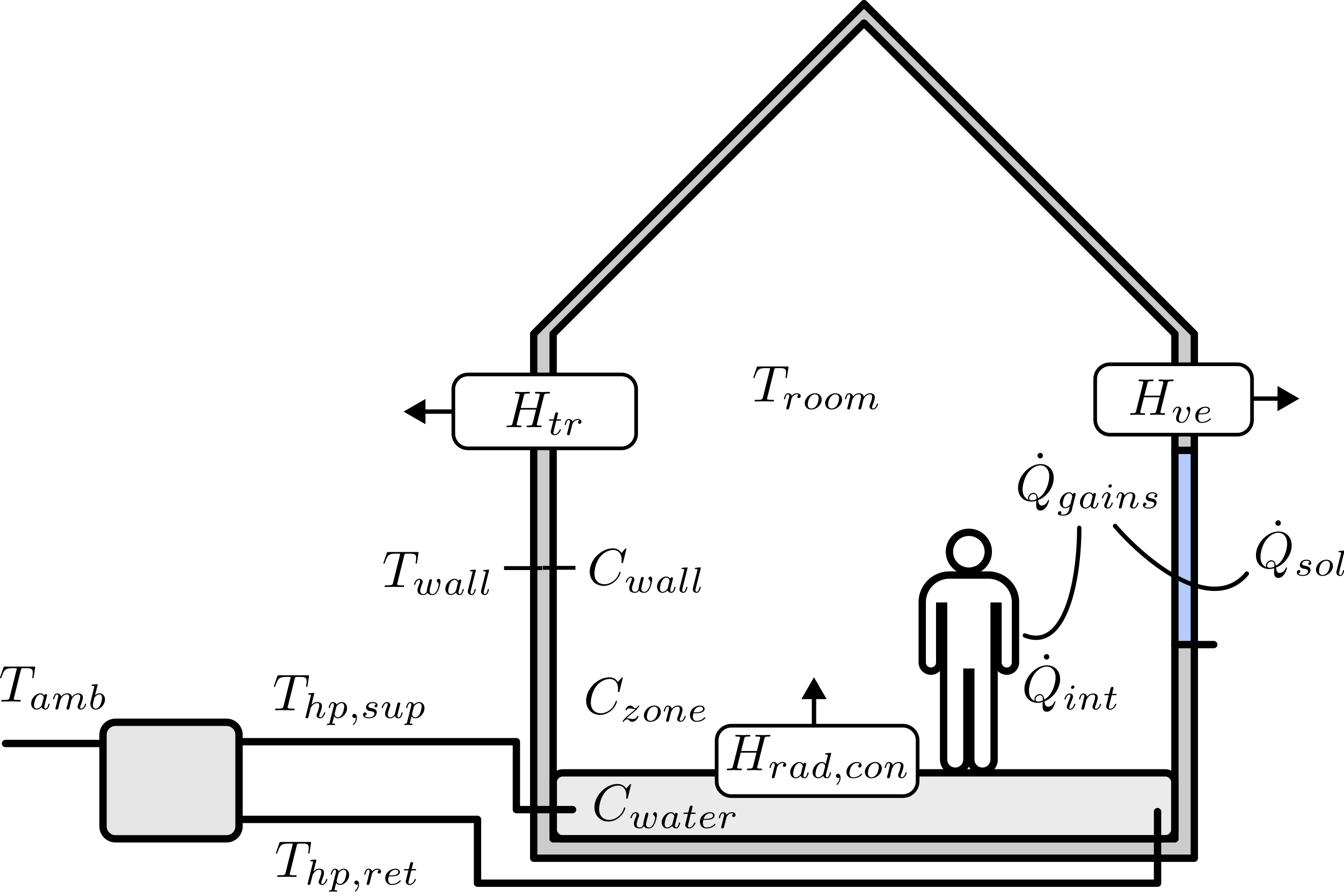}
    \caption{Example thermal building model}
    \label{fig:i4c-model}
\end{figure}

A thermal building model is a mathematical description of the thermal energy-related behavior of a building, including details about its structure, systems, usage, and location. Every model simplifies reality, often omitting certain phenomena, a principle particularly relevant to thermal behavior modeling in buildings, which involves complex interactions among internal conditions, external disturbances, and building materials. We adopt a single-zone model for simulating building thermal dynamics, striking a balance between detail and computational efficiency. 
\begin{table}[h]
    \resizebox{\columnwidth}{!}{ 
    \begin{tabular}{|l|l|}
        \hline
        \textbf{Symbol \& Unit} & \textbf{Description} \\
        \hline
        \multicolumn{2}{|l|}{\textbf{Input Parameters}} \\
        \hline
        $H_\mathrm{ve,tr}$ [\si{W/K}] & Heat transfer coefficient of transmission and ventilation \\
        $c_\mathrm{bldg}$ [\si{J/(m^2K)}] & Specific heat capacity of building \\
        $A_\mathrm{floor}$ [\si{m^2}] & Conditioned floor area \\
        $h_\mathrm{room}$ [\si{m}] & Average room height \\
        \hline
        \multicolumn{2}{|l|}{\textbf{Calculated Parameters}} \\
        \hline
        $C_\mathrm{bldg}$ [\si{J/K}] & Heat capacity of building envelope and thermal zone \\
        $C_\mathrm{water}$ [\si{J/K}] & Heat capacity of water in HVAC system \\
        $H_\mathrm{rad,con}$ [\si{W/K}] & Heat transfer coefficient of the heating system \\
        \hline
        \multicolumn{2}{|l|}{\textbf{Time-Varying Parameters}} \\
        \hline
        $T_\mathrm{amb}$ [\si{\degreeCelsius}] & Ambient temperature \\
        $Q_\mathrm{gains}$ [\si{W}] & Internal and solar heat gains \\
        \hline
        \multicolumn{2}{|l|}{\textbf{States}} \\
        \hline
        $T_\mathrm{room}$ [\si{\degreeCelsius}] & Temperature of the building \\
        $T_\mathrm{hp,ret}$ [\si{\degreeCelsius}] & Heat pump return flow temperature \\
        \hline
        \multicolumn{2}{|l|}{\textbf{Control}} \\
        \hline
        $T_\mathrm{hp,sup}$ [\si{\degreeCelsius}] & Heat pump supply temperature \\
        \hline
    \end{tabular}
    }
    \caption{Building model parameters}
    \label{tab:inputdata}
\end{table}
Our model focuses on a water-based heating system, where heat, transferred from water through radiators or underfloor heating, circulates back to the heat source at a reduced temperature, as depicted in Fig.~\ref{fig:i4c-model}. This return temperature, considered a state in our model, along with the net energy flow, dictates the indoor temperature changes. I4B also supports a more complex model that might include additional temperature nodes for walls and differentiates between transmission losses and thermal mass.

Exemplary, Table \ref{tab:inputdata} provides a definition of various building attributes and model quantities. For simplicity, we describe the formulas for a two-state model with the state vector~$x(t)=[T_\mathrm{room}, T_\mathrm{hp,ret}]$, capturing the thermodynamic interactions:
\begin{align}
	\dot{T}_\mathrm{room}& = 1/C_\mathrm{bldg} \cdot ( \dot{Q}_\mathrm{gain} + H_\mathrm{rad,con} \cdot (T_\mathrm{hp,ret} - T_\mathrm{room})\nonumber\\ & - H_\mathrm{ve,tr} \cdot (T_\mathrm{room}- T_\mathrm{amb}) ) 
\label{eq:dynamics0} \\
\dot{T}_\mathrm{hp,ret} &= 1/C_\mathrm{water} \cdot ( \dot{m}_\mathrm{hp} \cdot c_\mathrm{p,water} \cdot (T_\mathrm{hp,sup} - T_\mathrm{hp,ret}) \nonumber\\ &- H_\mathrm{rad,con} \cdot (T_\mathrm{hp,ret} - T_\mathrm{room}) ) 
\label{eq:dynamics}
\end{align}
\subsection{Heat pump model}\label{sec:hp}
The thermal power output of the heat pump, $\dot Q_\mathrm{hp}$, can be calculated using the supply and return temperatures ($T_\mathrm{hp,sup}$ and $T_\mathrm{hp,ret}$) and the typically constant mass flow rate $\dot m_\mathrm{hp}$ in the building's heat emission system. To determine the heat pump's efficiency, we consider that it uses mechanical energy to transfer heat from a lower to a higher temperature level—a process idealized by a reversible Carnot cycle, yielding a theoretical COP:
$\mathrm{COP}_\mathrm{th}=\frac{\dot{Q}_\mathrm{th}}{P_\mathrm{el}}=\frac{T_\mathrm{hp,sup}}{T_\mathrm{hp,sup}-T_\mathrm{amb} }$.
However, real heat pumps do not operate loss-free. The achievable COP of a heat pump is therefore smaller than the theoretical value by the exergetic efficiency $\eta_\mathrm{hp}$:
$\mathrm{COP} = \eta_\mathrm{hp}\mathrm{COP}_\mathrm{th}$.
Modern air-to-water heat pumps achieve
an efficiency of about 0.45. Rather than assuming a constant efficiency factor, the coefficient of performance (COP) is often modeled as a second-order polynomial function of the source and sink temperatures ($T_\mathrm{hp,sup}$ and $T_\mathrm{amb}$, respectively), with coefficients obtained from manufacturer data.

\subsection{Test cases and data}
The building model parameters defined in Tab.~\ref{tab:inputdata}, can be obtained from the open-source building typology tool TABULA~\cite{tabula}, providing information for a large number of different buildings in European countries.
The disturbance profiles for occupancy and appliances are derived according to standardized profiles for the corresponding building class. Solar gains are computed for each building using weather data and building information such as dimensioning of windows, orientation of the building, etc.

\section{Experiments}
\label{sec:experiments}
We consider the following research questions:
1).~How do controllers perform when sensor readings are corrupted noise on the temperature channels, i.e. noise robustness? 2).~How sensitive is MPC when prediction dynamics match the simulator vs.\ when inputs are noisy, compared to model‑free RL trained from the noisy observations (model mismatch)? 3).~What advantage does CSAC-LB bring by exploring near the comfort boundary without frequent violations (boundary behavior)?
\subsection{Experiment Setup}
\paragraph{Environment}
To evaluate constrained RL algorithms, we selected two real-world buildings: Building 1 is an older, poorly insulated structure with a water heat pump and ground collectors, leading to lower energy efficiency; Building 2 is a newer building utilizing an air-source heat pump. Simulations incorporate actual weather data and internal gains from occupants. To assess performance under noisy conditions, we add independent Gaussian sensor noise $\mathcal{N}(0, 0.5)$ in Kelvin to all temperature readings.
Our simulations use a three-state building model with 15-minute intervals per step. Controllers observe ambient temperature, room temperature, wall temperature, heat pump return water temperature, and energy contributions from alternative heat sources like solar radiation and internal gains. Their sole action is adjusting the heat pump's set temperature in °C, aiming to maintain a reference room temperature constraint of 20°C.
The objectives are minimizing electrical energy consumption (kWh) and keeping temperature above the reference. We regard a policy as comfort-compliant if its  average and yearly maximum temperature deviations remain below 0.1 K and 2.5 K, respectively.
\paragraph{Training Setup}
Heating systems present a unique challenge for constrained RL algorithms due to their data distribution: most real-world transitions cluster within a narrow range and rarely breach constraints. RL methods, starting without specific guidance, must learn strategies through trial and error. To encourage exploration across diverse states, we set episode lengths to 96 steps (one day) and reset the environment to random initial states after each episode, aiding the agent in navigating unsafe states. We evaluate the RL algorithms every 10 episodes over a full year and train them for 10,000 episodes/days. For MPC, we utilize the simulator's ground-truth dynamics for prediction. The temperature slack variable's weighting factor is set to .1 to balance comfort maintenance with energy cost minimization. We used the same hyperparameters for training the RL policies as in \cite{zhang2024constrained}.
\begin{figure}[t]
  \centering
  \includegraphics[width=\columnwidth]{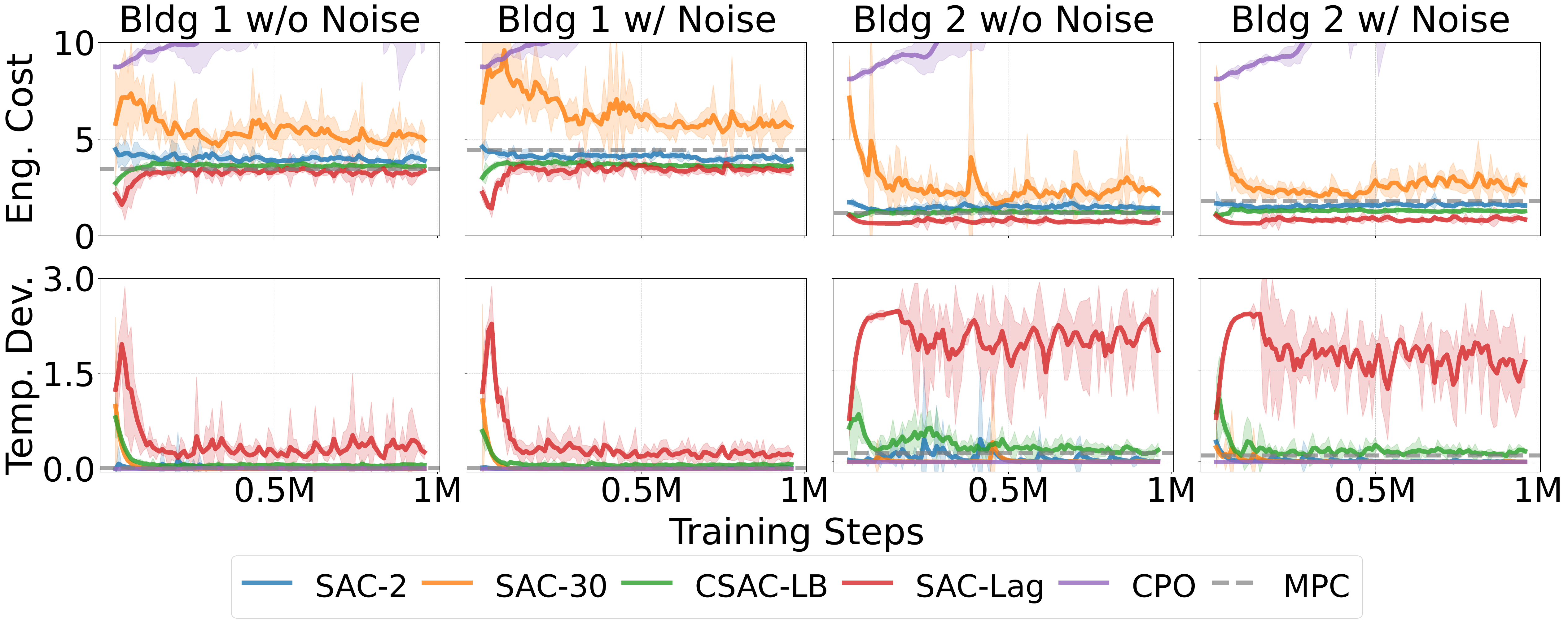}

  \caption{The mean and std of the evaluation results of different constrained RL algorithms obtained by averaging over 3 seeds. $x$-axis shows the number of training steps and $y$-axis on the top row shows the total amount of yearly electricity usage in kWh. The bottom row shows the averaged temperature deviation in Kelvin. MPC controller result is shown in grey dashed lines.}
  \label{fig:evaluation_curves}
  \vspace{-10pt}
\end{figure}

\paragraph{Baselines Algorithms}
We benchmark the following algorithms:~(a) \textbf{SAC}~\cite{DBLP:conf/icml/HaarnojaZAL18} with reward shaping: we add a penalty reward of -2/-30 when the constraint is violated, which is tuned by~\cite{DBLP:conf/nips/LuoM21}.~(b) \textbf{SAC-Lagrangian}~\cite{ray2019benchmarking}~(c) \textbf{CPO}~\cite{DBLP:conf/icml/AchiamHTA17}~(d) \textbf{CSAC-LB}~\cite{zhang2024constrained}~(e) \textbf{MPC}~\cite{Frison_2019}. ~(f) \textbf{Rule-based Controller}: A rule-based controller which only requires the ambient temperature of the building.
For MPC we use the simulator dynamics for prediction and the same weather and gains as RL.
\subsection{Experiment Results and Analysis}
\paragraph{Results}
Fig.~\ref{fig:evaluation_curves} compares the performance of the different RL algorithms during training to that of MPC. It reveals that CPO experiences training divergence; although it ensures satisfactory thermal comfort in most cases, its energy usage escalates drastically throughout training and fails to decrease, indicating unstable training. SAC-Lag displays unstable training patterns across all test environments, failing to meet expectations in both thermal comfort and energy efficiency. Both SAC variants secure adequate thermal comfort but at the cost of increased energy consumption, accompanied by noisy training trajectories characterized by significant fluctuations.

Regarding the MPC baseline, while it successfully maintains comfort across all scenarios (as seen in Tab.~\ref{table:kpis}), it does so at the cost of significantly higher energy consumption compared to the best-performing RL agents. For instance, in Building 2 with noise, MPC consumes 3427 kWh compared to 1312 kWh for CSAC-LB. This suggests that the MPC's tuning—specifically the weight on the slack variable—prioritizes constraint satisfaction too conservatively, leading to excessive energy use. In contrast, RL controllers use the same set of hyperparameters across both buildings, indicating greater robustness and transferability. Notably, all RL algorithms, except CPO, maintain comparable performance when subjected to noisy inputs. MPC also shows sensitivity to noise, with energy consumption rising by approximately 10-14\% in noisy environments, whereas CSAC-LB remains stable (e.g., 3652 kWh vs. 3608 kWh in Building 1). Among all methods, CSAC-LB balances thermal comfort and energy efficiency well, minimizing temperature deviations and energy use while maintaining strict adherence to the comfort constraints.

\begin{figure}[h]
    \centering
    \includegraphics[width=\columnwidth]{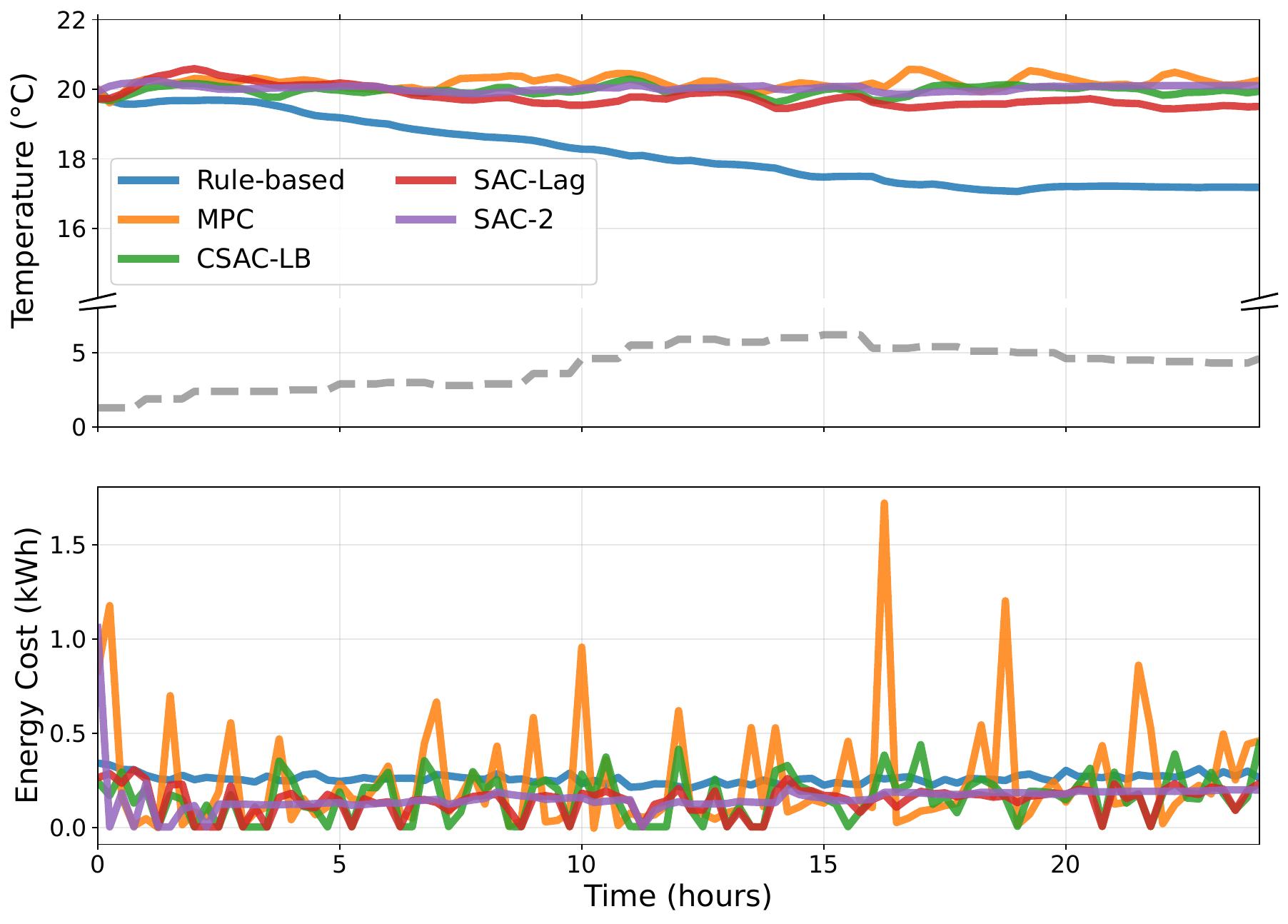}
    \caption{Illustration of ground truth room temperature achieved by running different controllers on Building 1 with added noise (top) and the energy cost of each controller at 15-min interval (bottom).}
    \label{fig:comparison_temperature}
\end{figure}

Fig.~\ref{fig:comparison_temperature} illustrates the ground-truth room temperature trajectories under injected sensor noise and the corresponding $15$\,min energy consumption for Building~1. CSAC-LB maintains a stable indoor temperature that tracks the comfort setpoint with minimal overshoot and rare boundary excursions, while exhibiting smooth actuation patterns in the energy cost curve. By contrast, SAC-Lag displays pronounced oscillations, frequently crossing into the unsafe comfort band. The MPC controller also shows undesirable behavior: while it successfully avoids comfort violations, it consistently overshoots the 20°C setpoint, often heating the room to 21°C. This conservative overheating likely explains its significantly higher energy consumption (5675 kWh vs. 3608 kWh in Tab.~\ref{table:kpis}). Furthermore, we observe rapid on/off switching of the heat pump (chattering) for the MPC; although not explicitly penalized in our objective, such chattering is undesirable in practice due to wear and reduced efficiency. Overall, CSAC-LB achieves tighter comfort regulation with smoother control effort under the same noisy observations.

\begin{figure}[h]
  \centering
  \subfloat[CSAC-LB]{%
    \includegraphics[width=0.32\columnwidth]{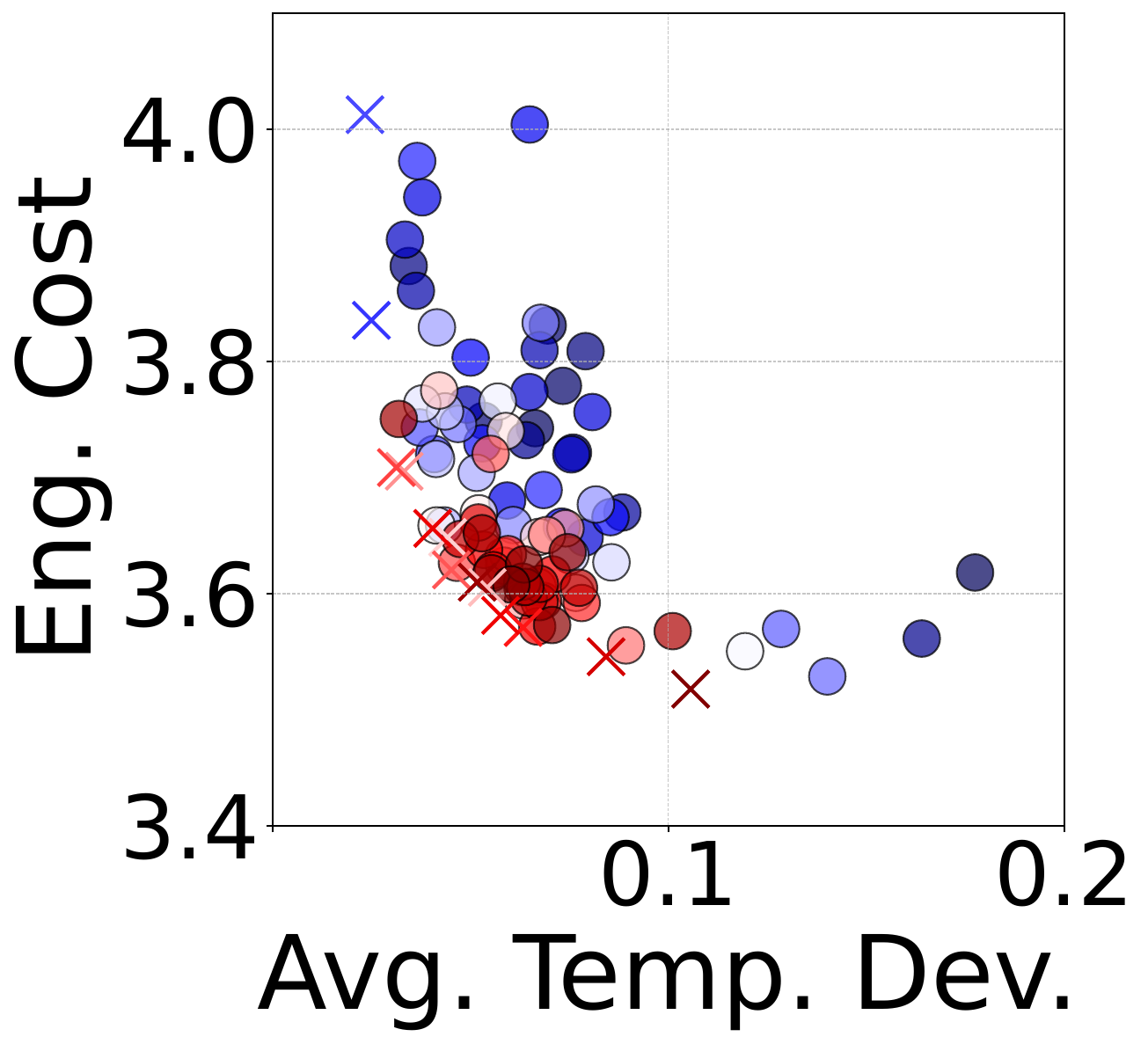}}%
  \hfill
  \subfloat[SAC-2]{%
    \includegraphics[width=0.27\columnwidth]{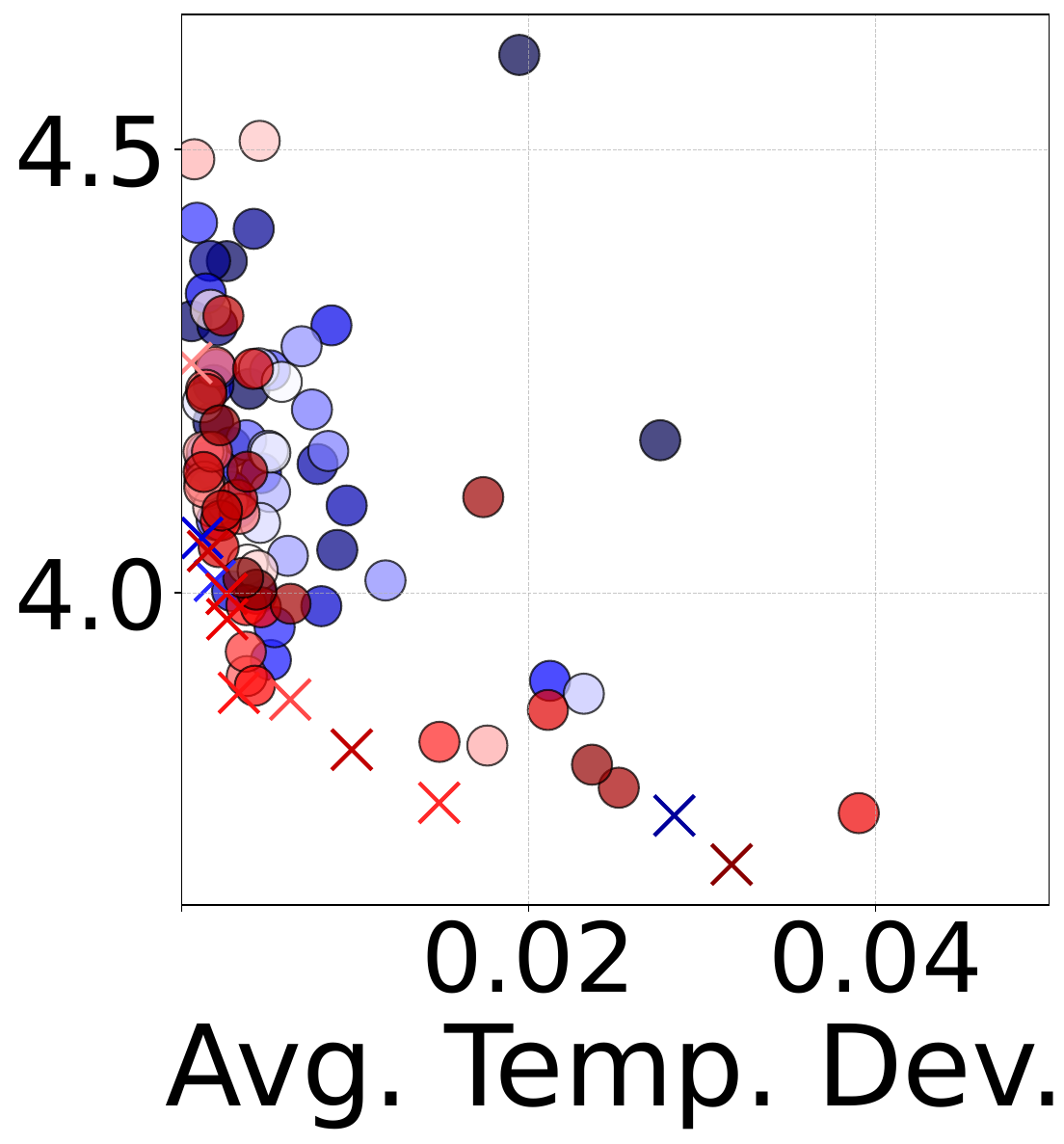}}%
  \hfill
  \subfloat[SAC-Lag]{%
    \includegraphics[width=0.36\columnwidth]{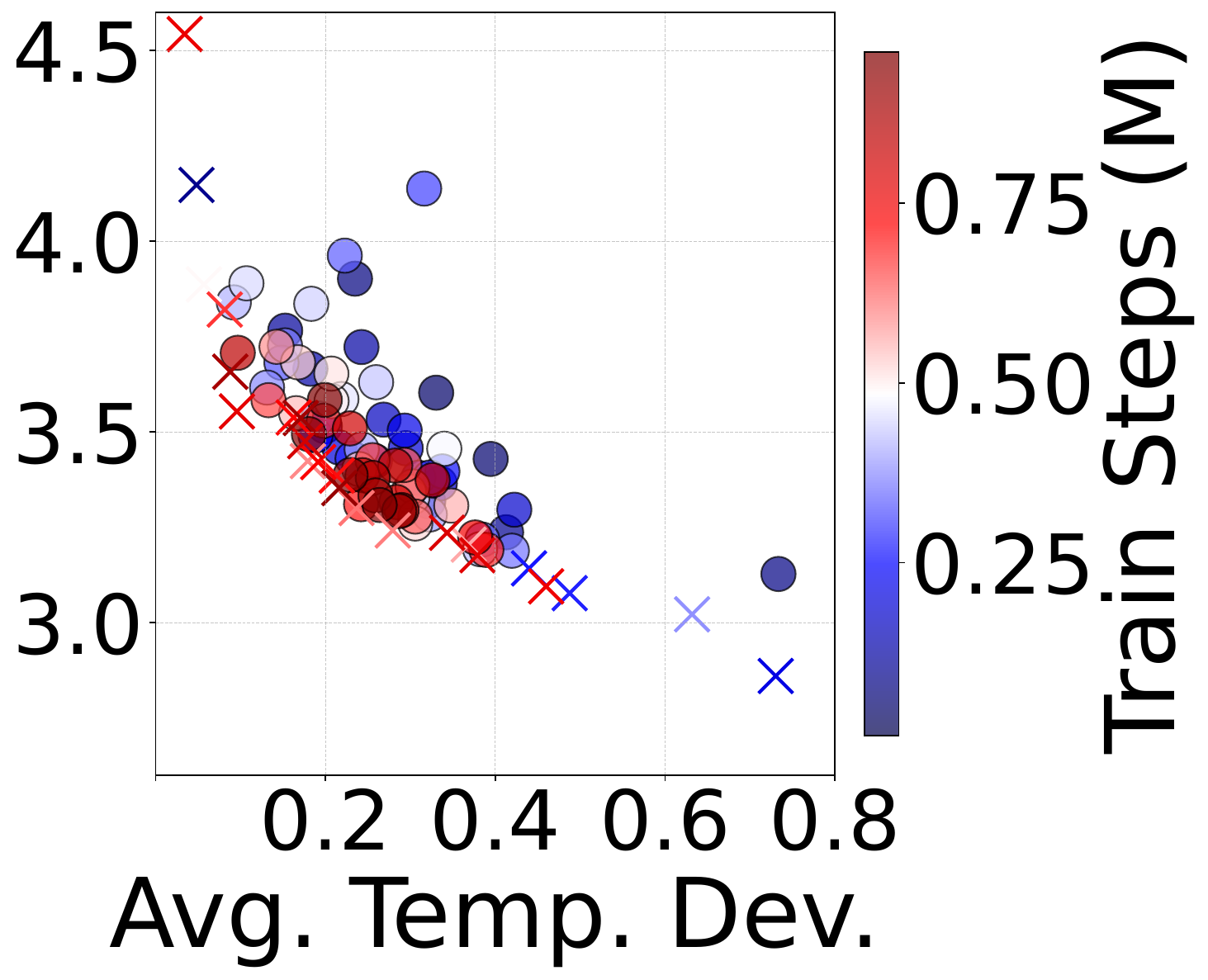}}%
  \caption{Illustration of the constrained RL algorithms evaluation results during training in Building 1 with noise. Each data point represents one evaluation episode and its color represents the number of training steps when the evaluation is performed. The $x$-axis is the average temperature deviation and the $y$-axis is the energy cost in kWh. Crosses mark evaluations on the Pareto front.}
  \label{fig:pareto}
\end{figure}

Considering performance during training, the example for Building 1 with noise, as illustrated in Fig.~\ref{fig:pareto}, shows CSAC-LB's exploration capabilities. CSAC-LB exhibits a more stable training trajectory, progressively advancing the energy–comfort Pareto frontier: later evaluations (red) concentrate near the comfort boundary with reduced energy. In contrast, SAC-Lag shows irregular, zig-zagging behaviors, leading to evaluations that deviate significantly from the Pareto line (highlighted by the diverse color scheme). Notably, CSAC-LB learns to operate consistently near the 20°C comfort bound (the 'safe margin') with minimal constraint violations, demonstrating its ability to learn from the most informative states without diverging. This performance demonstrates CSAC-LB's efficacy, which can be attributed to the advantages of using log barrier methods that effectively guide exploration, aligning with our rationale in Sec.~\ref{sec:csac_lb}.

\renewcommand{\arraystretch}{1.2} %
\begin{table*}[h]
\vspace{5pt}
\centering
\resizebox{0.95\textwidth}{!}{%
\begin{tabular}{|c|ccc|ccc|ccc|ccc|}
	\hline
	& \multicolumn{3}{c|}{Building 1}                             & \multicolumn{3}{c|}{Building 1 w. Noise}                              & \multicolumn{3}{c|}{Building 2}                             & \multicolumn{3}{c|}{Building 2 w. Noise}                           \\ \hline
	Algo       & \multicolumn{1}{c|}{E}  & \multicolumn{1}{c|}{$\bar{D}$} & $D_{\text{max}}$ & \multicolumn{1}{c|}{E}  & \multicolumn{1}{c|}{$\bar{D}$} & $D_{\text{max}}$ & \multicolumn{1}{c|}{E}  & \multicolumn{1}{c|}{$\bar{D}$} & $D_{\text{max}}$ & \multicolumn{1}{c|}{E}  & \multicolumn{1}{c|}{$\bar{D}$} & $D_{\text{max}}$ \\ \hline
	CPO        & \multicolumn{1}{c|}{19724} & \multicolumn{1}{c|}{yes} & 2.08\,(yes) & \multicolumn{1}{c|}{24269} & \multicolumn{1}{c|}{yes} & 0.06\,(yes)  & \multicolumn{1}{c|}{30900} & \multicolumn{1}{c|}{yes} & 0.00\,(yes)  & \multicolumn{1}{c|}{27564} & \multicolumn{1}{c|}{yes} & 0.00\,(yes) \\ \hline
	SAC-2      & \multicolumn{1}{c|}{3803}  & \multicolumn{1}{c|}{yes} & 1.13\,(yes)  & \multicolumn{1}{c|}{4017}  & \multicolumn{1}{c|}{yes} & 1.78\,(yes)  & \multicolumn{1}{c|}{1433}  & \multicolumn{1}{c|}{yes} & 0.15\,(yes)  & \multicolumn{1}{c|}{1569}  & \multicolumn{1}{c|}{yes} & 0.07\,(yes)  \\ \hline
	SAC-30     & \multicolumn{1}{c|}{4729}  & \multicolumn{1}{c|}{yes} & 0.16\,(yes)  & \multicolumn{1}{c|}{5542}  & \multicolumn{1}{c|}{yes} & 0.15\,(yes)  & \multicolumn{1}{c|}{1897}  & \multicolumn{1}{c|}{yes} & 0.06\,(yes)  & \multicolumn{1}{c|}{2588}  & \multicolumn{1}{c|}{yes} & 0.00\,(yes)  \\ \hline
	SAC-Lag    & \multicolumn{1}{c|}{3452}  & \multicolumn{1}{c|}{no} & 4.64\,(no)  & \multicolumn{1}{c|}{3584}  & \multicolumn{1}{c|}{no} & 3.47\,(no)  & \multicolumn{1}{c|}{852}   & \multicolumn{1}{c|}{no} & 5.70\,(no) & \multicolumn{1}{c|}{797}   & \multicolumn{1}{c|}{no} & 5.66\,(no)  \\ \hline
	CSAC-LB    & \multicolumn{1}{c|}{3652}  & \multicolumn{1}{c|}{yes} & 2.01\,(yes) & \multicolumn{1}{c|}{3608}  & \multicolumn{1}{c|}{yes} & 2.37\,(yes)  & \multicolumn{1}{c|}{1178}  & \multicolumn{1}{c|}{yes} & 1.33\,(yes)  & \multicolumn{1}{c|}{1312}  & \multicolumn{1}{c|}{yes} & 1.31\,(yes)  \\ \hline
	MPC        & \multicolumn{1}{c|}{5124}  & \multicolumn{1}{c|}{yes} & 0.54\,(yes)  & \multicolumn{1}{c|}{5675}  & \multicolumn{1}{c|}{yes} & 2.06\,(yes)  & \multicolumn{1}{c|}{3011}  & \multicolumn{1}{c|}{yes} & 1.26\,(yes)  & \multicolumn{1}{c|}{3427}  & \multicolumn{1}{c|}{yes} & 1.98\,(yes)  \\ \hline
	Rule & \multicolumn{1}{c|}{2901}  & \multicolumn{1}{c|}{no} & 3.21\,(no)  & \multicolumn{1}{c|}{2905}  & \multicolumn{1}{c|}{no} & 3.21\,(no)  & \multicolumn{1}{c|}{2128}  & \multicolumn{1}{c|}{yes} & 0.17\,(yes)  & \multicolumn{1}{c|}{2135}  & \multicolumn{1}{c|}{yes} & 0.18\,(yes)  \\ \hline
\end{tabular}%
}
\caption{Summary of KPIs regarding electrical energy consumption (E, in kWh) and average ($\bar{D}$) and maximum temperature deviation ($D_{\text{max}}$) in Kelvin for two different buildings with or without added observation noise. Requirements satisfied: yes/no.}
\label{table:kpis}
\end{table*}
\paragraph{Discussion}
Performance is always a trade-off between comfort (or safety) and energy efficiency. Based on the yearly energy consumption and comfort deviation metrics presented in Tab.~\ref{table:kpis}, only CSAC-LB reliably balances energy efficiency and comfort across all scenarios. CPO results in significantly higher energy consumption, whereas SAC-Lag exhibits unacceptable comfort deviations. While MPC satisfies the maximum deviation constraint ($D_{\text{max}}$) across all buildings, it is considerably less energy-efficient than CSAC-LB. For example, in Building 2, MPC consumes over double the energy of CSAC-LB (3011 kWh vs. 1178 kWh) to maintain similar comfort levels. Furthermore, MPC's energy consumption degrades under noisy observations (e.g., rising from 3011 to 3427 kWh in Building 2), highlighting its sensitivity to model-plant mismatch, whereas CSAC-LB remains robust. A likely explanation for SAC-Lag's instability is that with rare violations in heating scenarios and an underestimated cost critic, its Lagrange multiplier $\beta$ oscillates, yielding zig-zag trajectories away from the boundary. The barrier objective in CSAC-LB, by contrast, preserves a gradient near the boundary and stabilizes exploration.

\section{Conclusion}
In this study, we apply the recently developed constrained RL algorithm CSAC-LB to the critical problem of energy-efficient heat pump temperature control. Our goal is to operate the heat pump with low electricity consumption while reliably maintaining indoor temperature bounds during both training and operation. To achieve this, we introduce I4B, a lightweight, open-source simulation framework specifically designed for heat pump control in buildings. I4B provides extensive customization options and simulates various real-world building scenarios, thereby bridging the gap between the control and RL communities with its user-friendly interfaces.

Our evaluation of constrained RL algorithms demonstrates CSAC-LB's suitability for heat pump control. By leveraging a linear smoothed log barrier function and a double-Q network, CSAC-LB effectively addresses common underestimation issues and promotes boundary exploration, which is key for heating control tasks. Among the five recent RL algorithms tested, CSAC-LB stands out for its resilience to noisy data, adeptly balancing thermal comfort and energy efficiency. Considering model-plant mismatch in the form of noise, CSAC-LB remains robust and compares favourably to MPC.

\section{Acknowledgement}
This work is funded by the Deutsche Forschungsgemeinschaft (DFG, German Research Foundation) – Project-ID 499552394 – SFB 1597 and the German Federal Ministry for the Environment, Nature Conservation and Nuclear Safety (BMU) on the basis of a resolution of the German Bundestag as part of the ‘KI-Leuchtturm’ project ‘Intelligence for Cities’ (I4C). This project is also funded by the Deutsche Forschungsgemeinschaft (DFG, German Research Foundation) under grants 417962828, 428605208 and 539134284, through EFRE (FEIH\_2698644) and the state of Baden-Württemberg.

\begin{center}
\includegraphics[width=0.7\columnwidth]{images/BW_LOGO.png}

\vspace{2em}

\includegraphics[width=0.7\columnwidth]{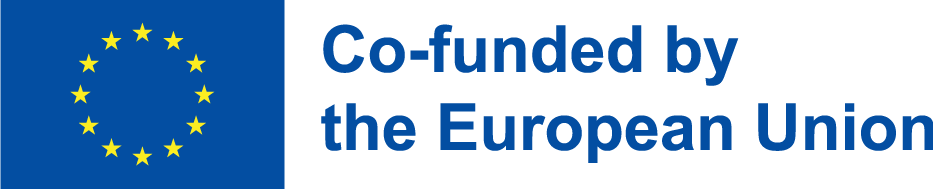}
\end{center}

\bibliographystyle{IEEEtran}
\bibliography{i4b}
\end{document}